\documentclass[10pt]{article} 
\usepackage[preprint]{tmlr}

\usepackage{amsmath,amsfonts,bm}

\def\eqref#1{equation~\ref{#1}}

\def\1{\bm{1}}

\DeclareMathAlphabet{\mathsfit}{\encodingdefault}{\sfdefault}{m}{sl}
\SetMathAlphabet{\mathsfit}{bold}{\encodingdefault}{\sfdefault}{bx}{n}

\usepackage{hyperref}
\usepackage{graphicx}
\usepackage{url}
\usepackage{dirtree}
\usepackage{booktabs}
\usepackage{colortbl} 
\usepackage{forest}
\usepackage{tikz}
\usepackage{adjustbox}
\usepackage{fontawesome5}
\usetikzlibrary{shadows}
\usetikzlibrary{positioning}
\usetikzlibrary{shadows.blur}

\usepackage{twemojis}

\newcommand{\codefont}[1]{\texttt{#1}}
\definecolor{folderbg}{RGB}{79, 129, 189}       
\definecolor{folderborder}{RGB}{61, 97, 141}    
\definecolor{darkgray}{gray}{0.3} 
\definecolor{lightgray}{gray}{0.8} 
\definecolor{slightdarkgray}{gray}{0.6}
\definecolor{lightblue}{RGB}{173,216,230}
\definecolor{darkerblue}{RGB}{135, 206, 250}
\definecolor{darkblue}{RGB}{0,0,139}
\newcommand{\errorBox}[1]{%
\begin{tikzpicture}
    \node[align=justify, line width=1.5pt, inner sep=3pt, text width=0.92\textwidth, anchor=center, font=\small, draw=red!5, fill=red!5, drop shadow, rounded corners, shift={(-1cm,-5cm)}] (errorDetails) at (0,0) {#1};
    \node[draw=red!50, circle, fill=red!50, minimum size=1cm, inner sep=0pt, anchor= west, drop shadow, shift={(-1cm,-0.5cm)}] at (errorDetails.north west) (errorIndicator) {};
    \node[font=\bfseries, scale=1.5, text=white] at (errorIndicator.center) {\twemoji{warning}}; 
\end{tikzpicture}
}

\newcommand{\tipBox}[1]{
\begin{tikzpicture}
    \node[align=justify, line width=1.5pt, inner sep=3pt,text width=0.92\textwidth, anchor=center, font=\small, draw=lightblue, fill=lightblue, drop shadow, rounded corners, shift={(-1cm,-5cm)}] (errorDetails) at (0,0) {#1};
    \node[draw=darkerblue, circle, fill=darkerblue, minimum size=1cm, inner sep=0pt, anchor= west,drop shadow, shift={(-1cm,-0.5cm)}] at (errorDetails.north west) (tipIndicator) {};
    \node[font=\bfseries, scale=1.5, text=white] at (tipIndicator.center) {\twemoji{rocket}}; 
\end{tikzpicture}
}

\newcommand{\implementationBox}[1]{
\begin{tikzpicture}
    \node[align=justify, line width=1.5pt, inner sep=3pt, text width=0.92\textwidth, anchor=center, font=\small, draw=lightgray, fill=lightgray, drop shadow, rounded corners, shift={(-1cm,-5cm)}] (errorDetails) at (1,0) {#1};
    \node[draw=slightdarkgray, circle, fill=slightdarkgray, minimum size=1cm, inner sep=0pt, anchor= west,drop shadow, shift={(-1cm,-0.5cm)}] at (errorDetails.north west) (tipIndicator) {};
    \node[font=\bfseries, scale=1.5, text=white] at (tipIndicator.center) {\twemoji{memo}};
\end{tikzpicture}
}

\def\Size{4pt}
\tikzset{
  folder/.pic={
    \filldraw[draw=folderborder,top color=folderbg!50,bottom color=folderbg]
      (-1.05*\Size,0.2\Size+5pt) rectangle ++(.75*\Size,-0.2\Size-5pt);  
    \filldraw[draw=folderborder,top color=folderbg!50,bottom color=folderbg]
      (-1.15*\Size,-\Size) rectangle (1.15*\Size,\Size);
  }
}

\title{A Partial Replication of MaskFormer in TensorFlow on TPUs for the TensorFlow Model Garden\includegraphics[width=0.03\textwidth]{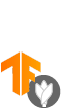}}
\author{\name Vishal Purohit \email purohitv@purdue.edu \\
\name Wenxin Jiang \email jiang784@purdue.edu \\
\name Akshath R. Ravikiran \email araviki@purdue.edu \\
\name James C. Davis \email davisjam@purdue.edu \\
\addr Elmore Family School of Electrical and Computer Engineering,\\
Purdue University, USA.
}

\def\month{MM}  
\def\year{YYYY}
\def\openreview{\url{https://openreview.net/forum?id=XXXX}} 

\begin{document}
\maketitle
\begin{abstract}
This paper undertakes the task of replicating the MaskFormer model — a universal image segmentation model — originally developed using the PyTorch framework, within the TensorFlow ecosystem, specifically optimized for execution on Tensor Processing Units (TPUs). Our implementation exploits the modular constructs available within the TensorFlow Model Garden (TFMG), encompassing elements such as the data loader, training orchestrator, and various architectural components, tailored and adapted to meet the specifications of the MaskFormer model.  We address key challenges encountered during the replication, non-convergence issues, slow training, adaptation of loss functions, and the integration of TPU-specific functionalities. We verify our reproduced implementation and present qualitative results on the COCO dataset. Although our implementation meets some of the objectives for end-to-end reproducibility, we encountered challenges in replicating the PyTorch version of MaskFormer in TensorFlow. This replication process is not straightforward and requires substantial engineering efforts. Specifically, it necessitates the customization of various components within the TFMG, alongside thorough verification and hyper-parameter tuning. Our implementation is available at \href{https://github.com/PurdueDualityLab/tf-maskformer/tree/main/official/projects/maskformer}{this link.}

\end{abstract}
\section{Introduction}
In recent years, machine learning has seen remarkable advancements, propelling forward the capabilities of AI systems across various domains. Despite these achievements, the machine learning community grapples with a persistent issue: reproducibility~\cite{Huston2018AIfacesReproducibilityCrisis}. The challenge of replicating existing machine learning models has emerged as a crucial hurdle, impeding the validation and further development of research findings. This reproducibility crisis underscores the discrepancies between theoretical models and their practical implementations, often leading to significant barriers to scientific progress within the field.

The importance of addressing the reproducibility crisis has been recognized by the machine learning community, prompting calls for comprehensive reproducibility reports~\cite{pineau_reproducible_2018, Pineau2020}. Such reports are vital for bridging the gap between research and practical application, ensuring that findings can be independently verified and built upon. In response to this call, our work focuses on the replication of MaskFormer~\cite{cheng2021maskformer}, a state-of-the-art image segmentation model originally developed in PyTorch, into the TensorFlow framework. This effort is not merely a technical exercise but a step towards establishing a more transparent, reproducible, and collaborative research environment.

Our paper documents the detailed process and challenges encountered in replicating MaskFormer in TensorFlow, offering insights into the nuances of cross-framework implementation. We delve into the specifics of adapting PyTorch-based methodologies to TensorFlow, highlighting the technical hurdles related to hardware specifications and the necessity of cross-framework model testing and verification. Through this replication study, we not only provide a roadmap for researchers aiming to undertake similar cross-framework projects but also shed light on broader issues of reproducibility. The lessons learned from this endeavor underscore the importance of detailed documentation, the adoption of standardized testing protocols, and the potential for future research to explore automated tools for model translation between frameworks. Our findings and experiences contribute to the ongoing dialogue on improving reproducibility in machine learning, proposing actionable directions for future work to tackle these systemic challenges.
\section{A Prelude to Universal Image Segmentation}
\begin{table*}[t]
    \centering
    \caption{A brief overview of TFMG modules.}
    \setlength\doublerulesep{1pt}
    \begin{adjustbox}{width=\textwidth}
    {\LARGE 
    \begin{tabular}{@{}ll@{}}
    \toprule 
    \textbf{TFMG Submodules}           & \textbf{Description} \\ 
    \midrule
    Official models                    & Designed and optimized models maintained by Google engineers.           \\
    Research models                    & Cutting-edge models from recent scientific literature.            \\
    Training experiment framework      & A declarative training environment that streamlines the process of configuring, training, and evaluating models.            \\
    Model training loop               & Declarative training mechanism that efficiently manages feeding data, updating model parameters, and computing gradients.           \\
    Specialized ML operations        & A set of TensorFlow operations for specific vision and natural language processing tasks.           \\
    \bottomrule
    \end{tabular}
    } 
    \end{adjustbox}
    \label{tab:overview_tfmg}
\end{table*}

\begin{figure}[]
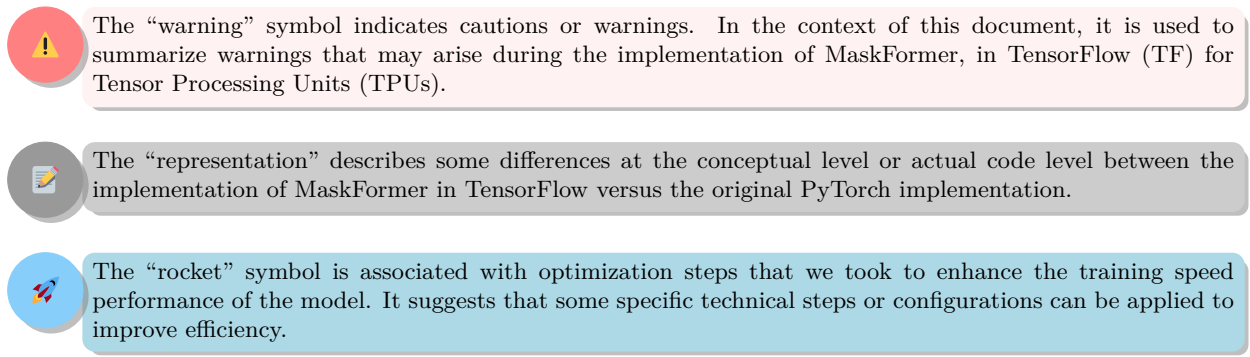

\centering
\errorBox{The ``warning'' symbol indicates cautions or warnings. In the context of this document, it is used to summarize warnings that may arise during the implementation of MaskFormer, in TensorFlow (TF) for Tensor Processing Units (TPUs).}

\implementationBox{The ``representation'' describes some differences at the conceptual level or actual code level between the implementation of MaskFormer in TensorFlow versus the original PyTorch implementation.}

\tipBox{The ``rocket'' symbol is associated with optimization steps that we took to enhance the training speed performance of the model. It suggests that some specific technical steps or configurations can be applied to improve efficiency.}
\caption{Description of the symbols used in this document.}
\label{fig:sym_legend}
\end{figure}
The field of image segmentation~\cite{unet, enc_dec_2018, Badrinarayanan2015SegNetAD, Chen2016DeepLabSI,zhou2019unetplusplus, pan_seg_2018, kirillov2023segany} categorizes pixels into various groups based on specific criteria, leading to the emergence of distinct segmentation tasks such as semantic, instance, and panoptic segmentation. Various segmentation tasks that arise due to specific grouping criteria are as follows:

\begin{itemize}
\vspace{-1.0em}
    \item\textbf{Semantic Segmentation:} Classifies each pixel in an image into a predefined category, without distinguishing between different objects of the same class.
\vspace{-1.0em}
    \item\textbf{Instance Segmentation:} Classifies and categorizes each pixel but also differentiates between individual objects of the same class.
\vspace{-1.0em}
    \item\textbf{Panoptic Segmentation:} Panoptic segmentation aims to offer a holistic view of the scene by merging the ``what'' and ``where'' aspects of both semantic and instance segmentation, treating every pixel in an image as part of either a ``stuff'' class (amorphous background elements like grass or sky) or a ``thing'' class (countable objects like people or vehicles)
\end{itemize}
\vspace{-0.5em}
Efforts to merge the diverse segmentation tasks—semantic, instance, and panoptic—into a unified task~\cite{cheng2021maskformer}. This approach eliminates the need for developing distinct architectures for each segmentation type, enabling a single model to deliver a detailed understanding of the scene. By consolidating these tasks, the model gains the versatility to analyze images comprehensively, providing both broad categorization and specific identification of individual objects within the same framework. This unified strategy aims to streamline the segmentation process, making it more efficient and adaptable to various applications. 
\begin{figure}[]
    \centering
    \includegraphics[width=\textwidth]{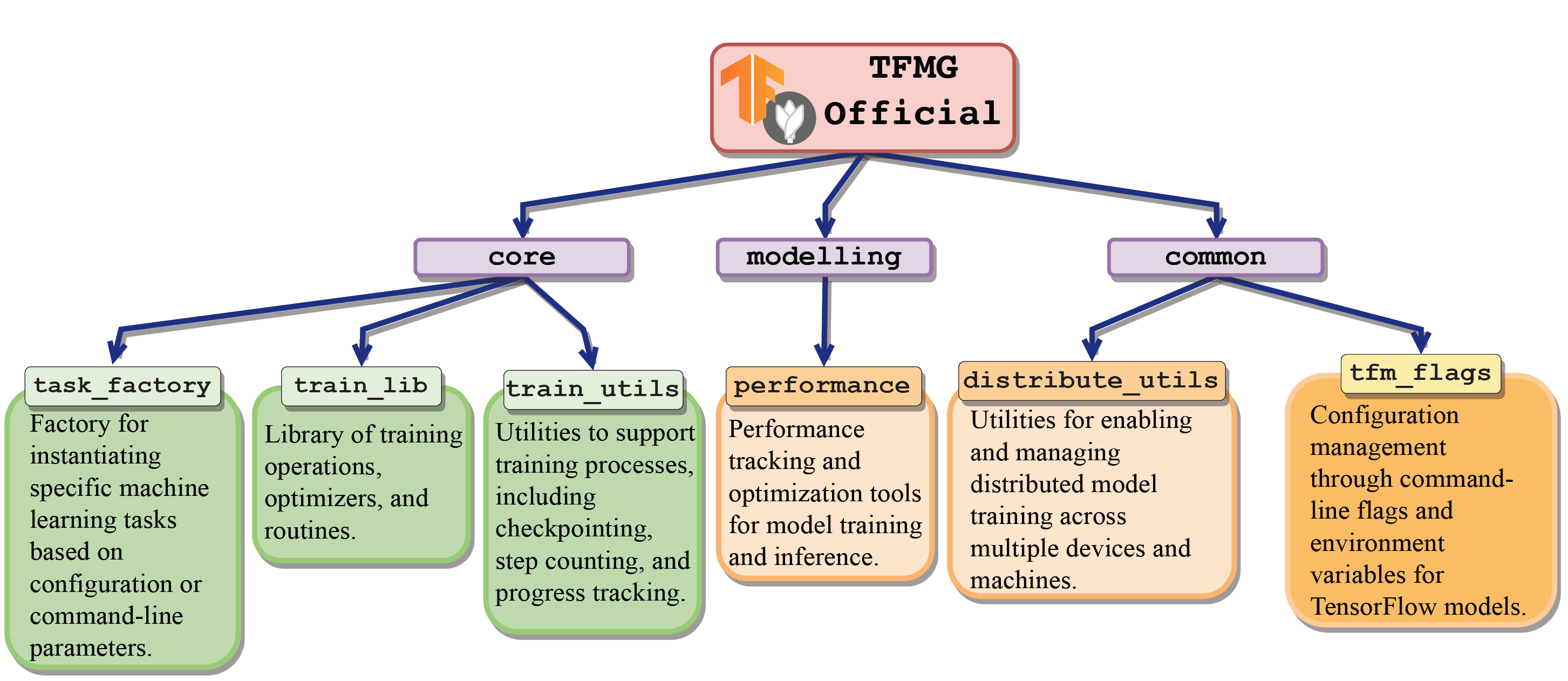}
    \caption{Overview of the TensorFlow Model Garden (TFMG) architecture. The diagram illustrates the modular design of the TFMG framework, highlighting the key components such as {\fontfamily{qcr}\selectfont `task\_factory'} for initializing machine learning tasks,
     {\fontfamily{qcr}\selectfont `train\_lib'} for training operations, {\fontfamily{qcr}\selectfont`train\_utils'} for training utilities, {\fontfamily{qcr}\selectfont`core'} for central functionalities, {\fontfamily{qcr}\selectfont`modeling'} for model definitions, {\fontfamily{qcr}\selectfont`performance'} for tracking and optimization, {\fontfamily{qcr}\selectfont`distribute\_utils'} for distributed training support, and {\fontfamily{qcr}\selectfont`tfm\_flags'} for configuration management. Arrows indicate the direction of dependencies between modules.
     }
    \label{fig:tfmg_train}
\end{figure}
The advent of universal architectures, as exemplified by DETR~\cite{carion2020end}, has revolutionized the field of image segmentation, demonstrating that mask classification architectures equipped with an end-to-end set prediction objective possess the versatility required for any segmentation task. Building on this foundation, MaskFormer further illustrates the efficacy of DETR-based mask classification, not only excelling in panoptic segmentation but also setting new benchmarks in semantic segmentation performance. This progression underscores a significant shift towards more generalized and efficient frameworks for tackling the diverse challenges of image segmentation. In this work, we reproduce the MaskFormer architecture in TensorFlow with inference and training supported on TPUs.

\section{Overview of TensorFlow Model Garden Library}

The TensorFlow Model Garden (TFMG)~\cite{tensorflowmodelgarden2020} is a comprehensive suite of state-of-the-art machine learning models, particularly focused on vision and natural language processing (NLP). It offers both official models maintained by Google engineers and research models from academic papers, along with tools for easy configuration and deployment on standard datasets. 
\begin{figure}[]
    \centering
    \includegraphics[width=\textwidth]{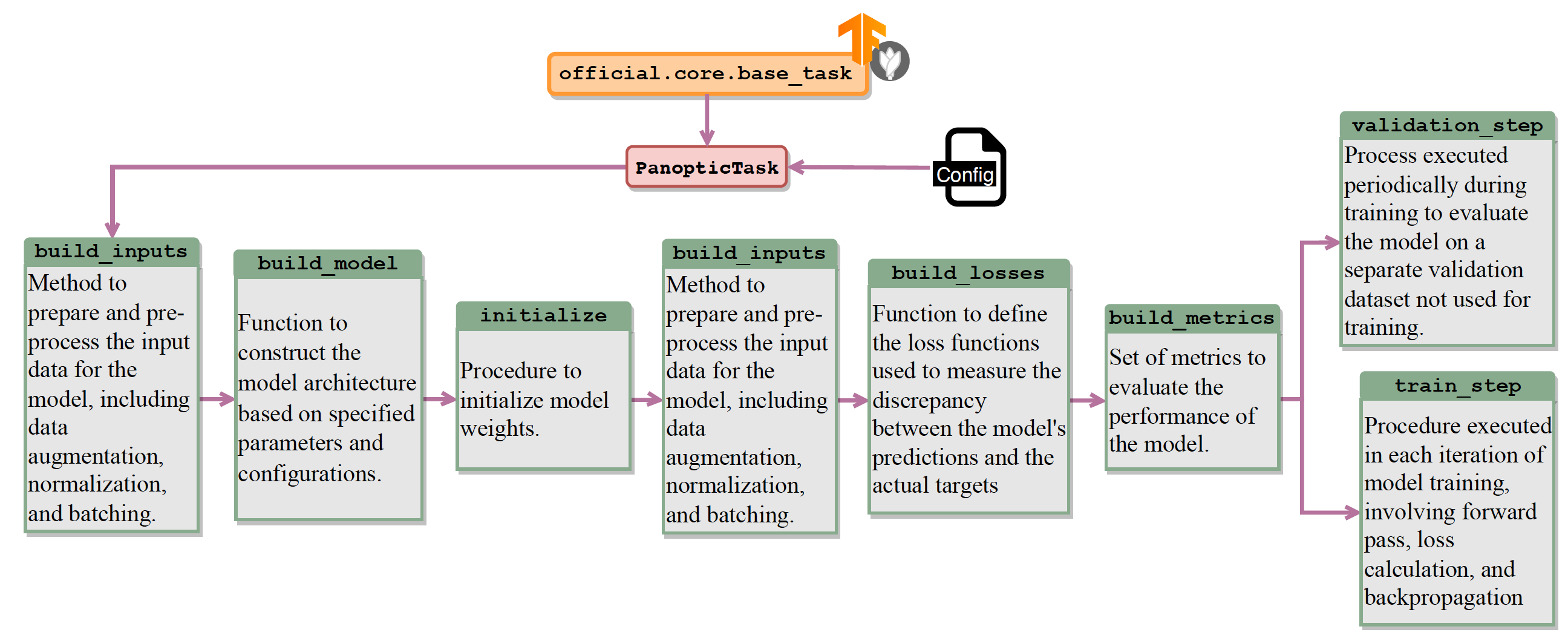}
    \caption{Illustration the workflow components of a machine learning task, centralized around {\fontfamily{qcr}\selectfont`PanopticTask'}. It includes the configuration, model building, initialization, and input preparation stages, followed by training and validation steps. {\fontfamily{qcr}\selectfont`build\_losses'} and {\fontfamily{qcr}\selectfont`build\_metrics'} are essential for calculating model performance during training.}
    \label{fig:panoptic_task}
\end{figure}
In this section, we highlight key TFMG components instrumental in developing the MaskFormer model in TensorFlow. The Table \ref{tab:overview_tfmg} provides an organized summary of the submodules in TFMG. These components range from well-maintained official models to cutting-edge research models and include a streamlined training environment as well as specialized operations tailored for vision and NLP tasks. This modular overview encapsulates the diverse utilities offered by TFMG for model development. The MaskFormer model's development harnesses the {\fontfamily{qcr}\selectfont official} module of TFMG, which encompasses a suite of submodules crafted for general model development. These submodules provide a foundation for tasks such as initializing specific machine learning tasks, managing training operations, optimizing model performance, and configuring distributed model training across multiple devices. A detailed visualization of the submodules, along with their specified functionalities within the TFMG {\fontfamily{qcr}\selectfont official} module is shown in Figure \ref{fig:tfmg_train}.

Typically, the development of a model utilizing the TFMG follows a structured sequence of steps. Following is the sequence of steps which are described in detail in subsequent sections:
\begin{enumerate}
    \item \textbf{Project Structure Organization:} Every project developed using TFMG requires following a well-defined structure to ensure the seamless integration of various components and to facilitate efficient development workflows. Specifically for MaskFormer, we follow the following directory structure. 
\begin{forest}
  for tree={
    font=\ttfamily,
    grow'=0,
    child anchor=west,
    parent anchor=south,
    anchor=west,
    calign=first,
    inner xsep=7pt,
    edge path={
      \noexpand\path [draw, \forestoption{edge}]
      (!u.south west) +(7.5pt,0) |- (.child anchor) pic {folder} \forestoption{edge label};
    },
    before typesetting nodes={
      if n=1
        {insert before={[,phantom]}}
        {}
    },
    fit=band,
    before computing xy={l=15pt},
  }  
[TFMG official
[projects - A high-level project directory for all the TFMG projects
  [maskformer - MaskFormer project directory.
  [configs - Contains model parameters and training settings.]
  [data - Contains scripts for creation of TFRecords.]
  [dataloaders - Contains data reader and parser for dataloader.]
  [losses - Contains implementation of loss functions]
  [modeling - Architecture implementation.]
  [tasks - Contains MaskFormer task that coordinates training and testing.]
  [utils - Contains utility functions.]
  ]
  [...]
  ]
]
\end{forest}
    \item \textbf{Compute and Environment:} We set up the training environment using Google Cloud Platform (GCP)  which offers scalable and versatile computing resources. Our implementation of MaskFormer has been designed to support training on both GPU (Graphics Processing Unit) and TPU (Tensor Processing Unit) platforms with the help of \codefont{ official.common.distribute\_utils}. The configuration of the distributed training is all handled by the training driver script described in step 5.
    \item \textbf{Task Specification:} The modular architecture of TFMG is designed for the seamless addition of new models through precise task specification, which is typically encapsulated in a script named \codefont{<task\_name>.py}, located within the \codefont{tasks} directory. At a higher level, these task specifications play a critical role in the model training process. They are utilized by the training driver script to initialize various critical aspects of model training, including setting up the training and validation processes and defining the metrics to evaluate model performance. In the specific case of the MaskFormer model, customization within our framework is achieved by extending the \codefont{base\_task} class, which is sourced from \codefont{official.core}. By overriding several of its member functions, we tailor the underlying implementation to meet the unique requirements of the MaskFormer model. An illustration of the modified member functions of the base class for MaskFormer with their corresponding functionalities is shown in Figure \ref{fig:panoptic_task}. The implementation details of these functions are delineated in subsequent sections. 
    \item \textbf{Configuration Specification:} This step involves detailing the parameters related to training and model configuration, including the hyperparameters critical for model performance optimization. This process is facilitated through the extension of the  \codefont{official.core.config\_definitions.TaskConfig} class, which acts as a container for all requisite model parameters.  The organization of configuration parameters is arranged, adhering to their functional relevance. This is achieved by grouping them under distinct classes derived from \codefont{official.modeling.hyperparams}. Some of the systematically defined hyperparameter classes  for MaskFormer include:
    \begin{itemize}
        \item \codefont{Parser:} Specifies the parameters for preprocessing input data, including output size, scale, and aspect ratio range.
        \item \codefont{DataConfig:} Outlines the settings for input data handling, such as paths to training and validation datasets, batch sizes, and data augmentation strategies. 
        \item \codefont{MaskFormer:} Details the structural specifics of the MaskFormer model, including the number of queries, hidden size, and the backbone used.
        \item \codefont{Losses:} Enumerates the different losses used during training, including class offset, background class weight, and L2 weight decay. 
        \item \codefont{PanopticQuality:} A custom evaluator configured to assess the panoptic quality (PQ) of our model's predictions
    \end{itemize}
    Note that these configs can be overridden by the user via the command line or by passing a \codefont{yaml} file to the training command.
    \item \textbf{Training Driver Script:} (\codefont{train.py}) The script serves as a training driver for the project. It handles the setup and execution of model training and evaluation across different computing environments, including CPU, GPU, and TPU. Key steps include parsing configuration specified in step 4, setting up the appropriate TensorFlow distribution strategy based on the computing environment (step 2), initializing the model and task based on provided configurations (step 4), and running the training or evaluation process (step 3). The TFMG project uses Orbit Trainer as a lightweight and flexible training and evaluation loop. Orbit simplifies the creation of custom training loops, reducing the boilerplate code. It supports both eager execution and graph execution modes.
\end{enumerate}

\section{Architecting MaskFormer in TensorFlow for TPUs}
In this section, we provide a concise overview of the data-loader, architecture, and loss functions that are key components critical to the reproducibility of MaskFormer. 

\begin{figure*}[t]
    \centering
    \includegraphics[width=\textwidth]{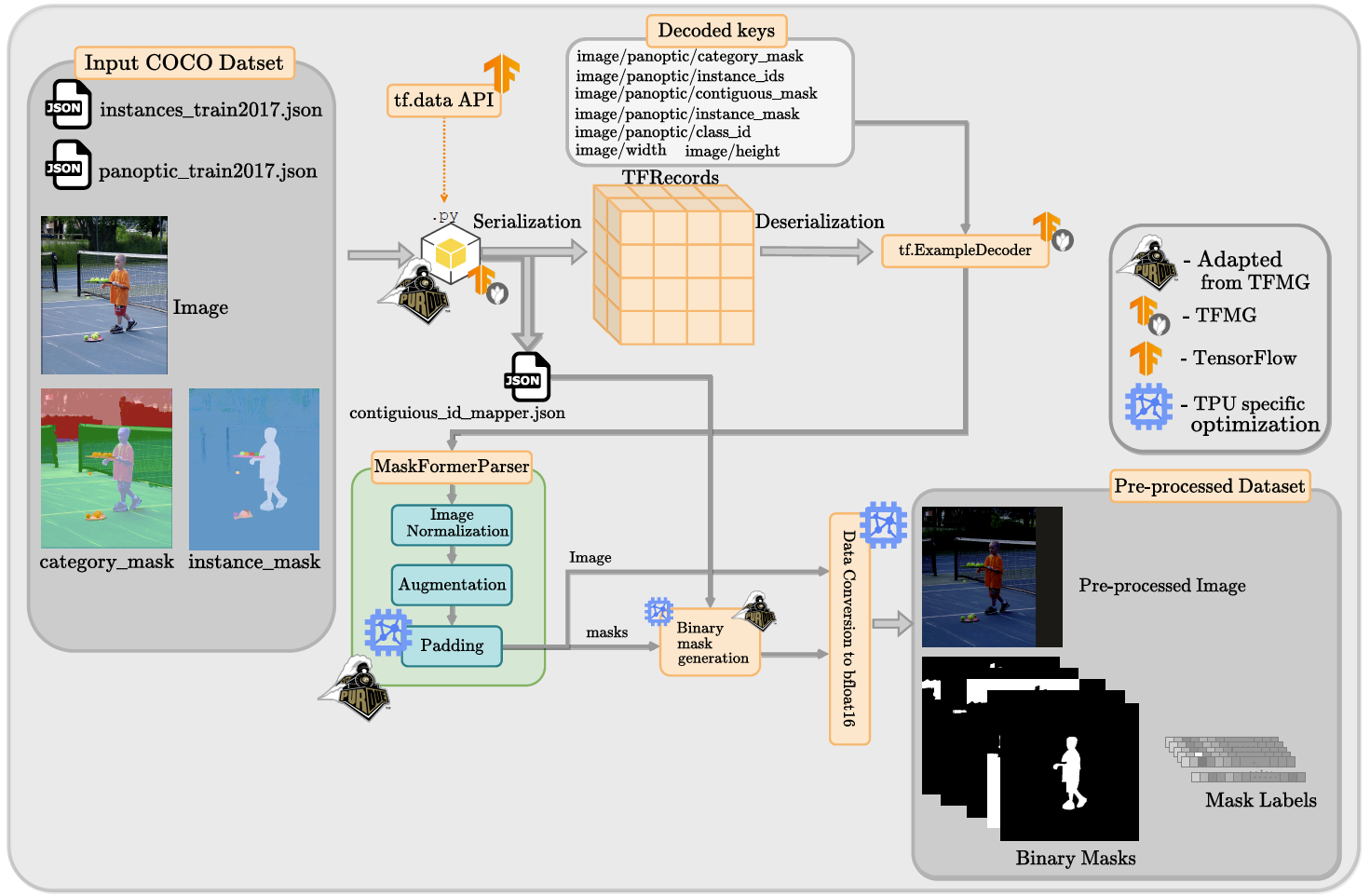}
    \vspace{-2.5em}
    \caption{Illustration of the data pre-processing pipeline for panoptic segmentation task, specifically for training MaskFormer model with `COCO' dataset. The process begins with the raw image and annotation files (JSON format), which are serialized into TFRecord format, a TensorFlow-specific binary storage format. Subsequently, these TFRecords are deserialized, and the data is passed through a series of preprocessing steps including normalization, augmentation, and padding to ensure uniformity of the input data. The binary mask generation and the label mapping are integral to preparing segmentation tasks. The output is then converted to a format compatible with the model’s data loader, which feeds the processed data into the training pipeline. (zoom in for a better viewing experience)}
    \vspace{-0.5em}
    \label{fig:maskfomer_data}
\end{figure*}
\subsection{Dataset \& Data loader}

\subsubsection{Dataset}\label{subsec:dataset}
In the context of our reproducibility project focused on the MaskFormer architecture, we chose to employ the COCO (Common Objects in Context)~\cite{cocodataset} dataset as the primary benchmark for training and evaluation. As a first step towards data preparation, we need to create the TFRecords, which is a binary file format used by TensorFlow for storing data. It supports the inclusion of various types of annotations such as detection bounding boxes, instance segmentation masks (optionally encoded as PNG images), and textual captions. For the creation of TFRecords of the COCO dataset, a standardized script is made available from the TFMG Vision package. Specifically, we adapt the base implementation \codefont{create\_coco\_tf\_record.py} made available at \codefont{official > vision > data}.

The MaskFormer architecture differs from existing panoptic segmentation models available on the TFMG by its unique requirement for per-object binary masks which have their corresponding class labels for loss calculation. In order to facilitate this we propose following to existing base implementation of  \codefont{create\_coco\_tf\_record.py} as follows:
\begin{itemize}
    \item \textbf{Conversion of non-contiguous coco class ids into contiguous :} By default in the COCO dataset, class IDs are non-contiguous, meaning there are gaps in the sequence of IDs used to label object categories. For our implementation, we adopted an approach to manage the conversion between non-contiguous (original) class IDs from the COCO dataset and contiguous IDs. We maintain a mapping between non-contiguous (original class ids) and the contiguous ids facilitated by a mapping stored in \codefont{contiguous\_id\_mapper.json}. This JSON mapping file acts as a lookup table, where each original class ID from the COCO dataset is associated with a new, contiguous ID. By utilizing this mapping, we can easily convert the non-contiguous class IDs present in the `category\_masks' for the COCO dataset to a `contiguous\_mask'. Further, the `contiguous\_mask' is used by the binary mask generation step in the dataloder (described in the next section). An illustration of TFRecord creation is shown in Figure \ref{fig:maskfomer_data}.
    \end{itemize}
    The final output of the TFRecord creation step is a set of binary files also known as shards. Each shard contains a subset of the dataset, formatted as TFRecord entries that TensorFlow can efficiently load. In TensorFlow's TFRecord format, data is serialized into a string format and stored in a key-value pair within a \codefont{tf.train.Example} message. The keys in this context are used to specify the type of data stored, allowing each data element to be uniquely identified and correctly deserialized during the data loading process. The set of keys used by our implementation is shown in Figure \ref{fig:maskfomer_data}.

    \tipBox{Ideally, generating binary masks during the creation of TFRecords is desirable, as it streamlines the data loading process by eliminating additional preprocessing steps. This approach integrates the generation of per-instance binary masks directly into the dataset preparation phase, ensuring that the model training process can directly consume the data without the need for on-the-fly mask generation. However, this method comes with a significant drawback: the increased size of TFRecord files. When training models in a distributed environment, such as on GCP, the enhanced size of TFRecords can become a critical bottleneck. The larger file sizes demand more bandwidth for data transfer and increase the time required for data loading, potentially slowing down the overall training performance. Additionally, TFRecord shards must be approximately equal in size to ensure that each worker has a balanced workload. Uneven shard sizes can result in some workers idling while waiting for others to complete their tasks, leading to inefficient resource utilization.}

\subsubsection{Dataloader}
The data pipeline for training MaskFormer consists of several stages. Our work reuses some of the stages of the data pipeline directly from pre-existing TFMG modules and customizes only the input reader and parser. The stages of the data pipeline are briefly described below along with our proposed modifications. 

    \textbf{Input Decoder and Parser:} The input decoder serves as the primary interface with the dataset, responsible for fetching the raw data. We integrate and adopt the \codefont{input\_reader.py} from \codefont{official > vision > dataloaders} directory of TFMG.

    \textbf{Decoder:} We customize the base \codefont{TFExampleDecoder} class from \codefont{official.vison.dataloaders.tf\_example\_decoder}. 
    This customization involves augmenting the class's initialization process to incorporate additional keys that are specific to the TFRecords generated for our TFRecords as explained in Section \ref{subsec:dataset}. By introducing these new keys during the class's instantiation, we ensure that our decoder is equipped to recognize and appropriately process these specialized data fields. The new keys are used by the \codefont{decode} member function. 
    
    \textbf{Parser:} Once the data is decoded from the decoder the parser handles all the pre-processing operations. We inherit the parent parser from \codefont{official.vision.dataloaders} and override required member functions. Firstly the role of the parser can be abstracted into two functional parts:
    \begin{itemize}
        \item \textbf{Image and Mask Processing :} Normalizes images, resize images, and masks to target dimensions and potentially applies data augmentation techniques like random horizontal flipping or cropping based on the training configuration.  Notably, the implementation of random cropping is executed with a certain probability $p$ and the cropped images' shortest side is resized to one of the predetermined sizes in the set [400px, 500px, 600px]. Further, we pad the cropped image and mask with zeros. Note that we explicitly ensure the padded region is ignored during the mask loss calculation step.
    \end{itemize}
    \implementationBox{Our implementation of the MaskFormer dataloader in TensorFlow presents significant deviations from its PyTorch counterpart, primarily due to the additional padding requirements to guarantee a consistent output size from the parser. Unlike the PyTorch dataloader, which does not enforce a uniform size for the dataloader's output, our approach incorporates explicit resizing and padding strategies. This adjustment is necessitated by the operational constraints of Tensor Processing Units (TPUs). TPUs, being highly optimized for parallel computations, require inputs to be of a uniform size to maximize computational efficiency and throughput. Consequently, our TensorFlow-based dataloader is specifically designed to meet these hardware requirements by ensuring that every output batch conforms to a predetermined size, thereby facilitating seamless processing on TPU architectures for optimal performance.}

    \tipBox{To optimize the efficiency and performance of TensorFlow input pipelines, our approach is grounded in the best practices outlined in the TensorFlow Core guide. These methodologies, which are intrinsically supported by the TFMG codebase, serve as a foundation for our data-handling strategies. Among these practices, a notable point of consideration is the application of the \codefont{cache} mechanism. Our experience indicates that while caching can substantially reduce the time spent on reading data from disk in subsequent epochs, thereby accelerating the training process, its effectiveness is contingent upon the size of the dataset. Specifically, we observed that caching is advantageous when the dataset is sufficiently small to fit into memory without causing resource constraints. However, for larger datasets, the attempt to cache the entire dataset often leads to memory overflow issues.}
    
    \begin{itemize}
        \item \textbf{Label Preparation :} The primary aim of the Label Preparation step is to process the ``contiguous\_mask" annotations and generate individual masks for each object instance. These masks are then associated with their corresponding class IDs. The end-to-end process of data preparation is shown in Figure \ref{fig:maskfomer_data}.
    \end{itemize}

    \errorBox{\textcolor{red}{\textbf{Error :} \codefont{tensorflow.python.framework.errors\_impl.UnavailableError: Socket closed}}\\
    \textbf{Context :} We encountered this error while debugging the data loader. \\
        \textbf{Description :} This error suggests that a network socket, which is necessary for the communication between different parts of your TensorFlow application or between your application and a TensorFlow service, was unexpectedly closed.\\
        \textbf{Solution :} While the error superficially points towards network communication issues, it may sometimes lead to misdirection during troubleshooting. Our extensive debugging efforts revealed that the root cause of the problem was not directly related to network connectivity. Instead, it stemmed from the utilization of the \codefont{tf.unique} operation within the preprocessing phase of our pipeline. This operation, we discovered, is incompatible with XLA (Accelerated Linear Algebra), a crucial performance optimization component of TensorFlow designed to accelerate linear algebra computations.}
    
\begin{figure}[]
    \centering
    \includegraphics[width=\textwidth]{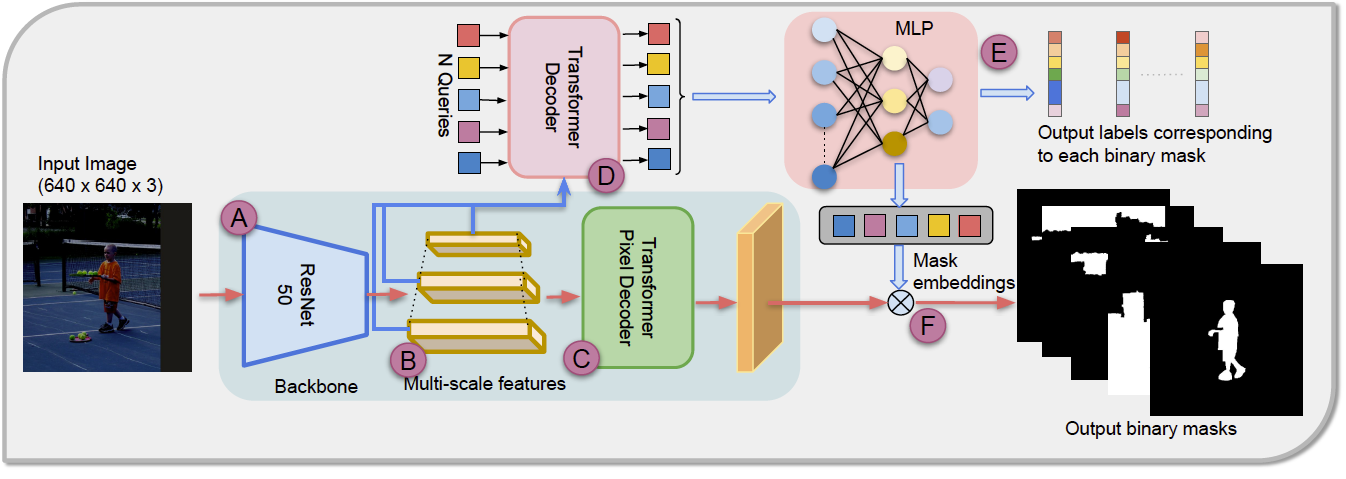}
    \caption{Illustration of MaskFormer architecture (reproduced from ~\cite{cheng2021maskformer}.). The process begins with an input image (A), which is processed by a ResNet-50 backbone to extract feature maps. Multi-scale features (B) are generated from the backbone and then decoded by a Pixel Decoder (C) to create refined feature maps. These are fed into a Transformer Decoder (D), along with a set of learned queries that interact with the feature maps to generate mask embeddings (F). An MLP head (E) processes the mask embedding to produce the final segmentation output, which consists of binary masks for each query, along with their corresponding output labels, indicating the presence of specific objects or regions within the input image.}
    \label{fig:maskfomer_arch_overview}
\end{figure}
\begin{figure}[t]
    \centering
    \includegraphics[width=\textwidth]{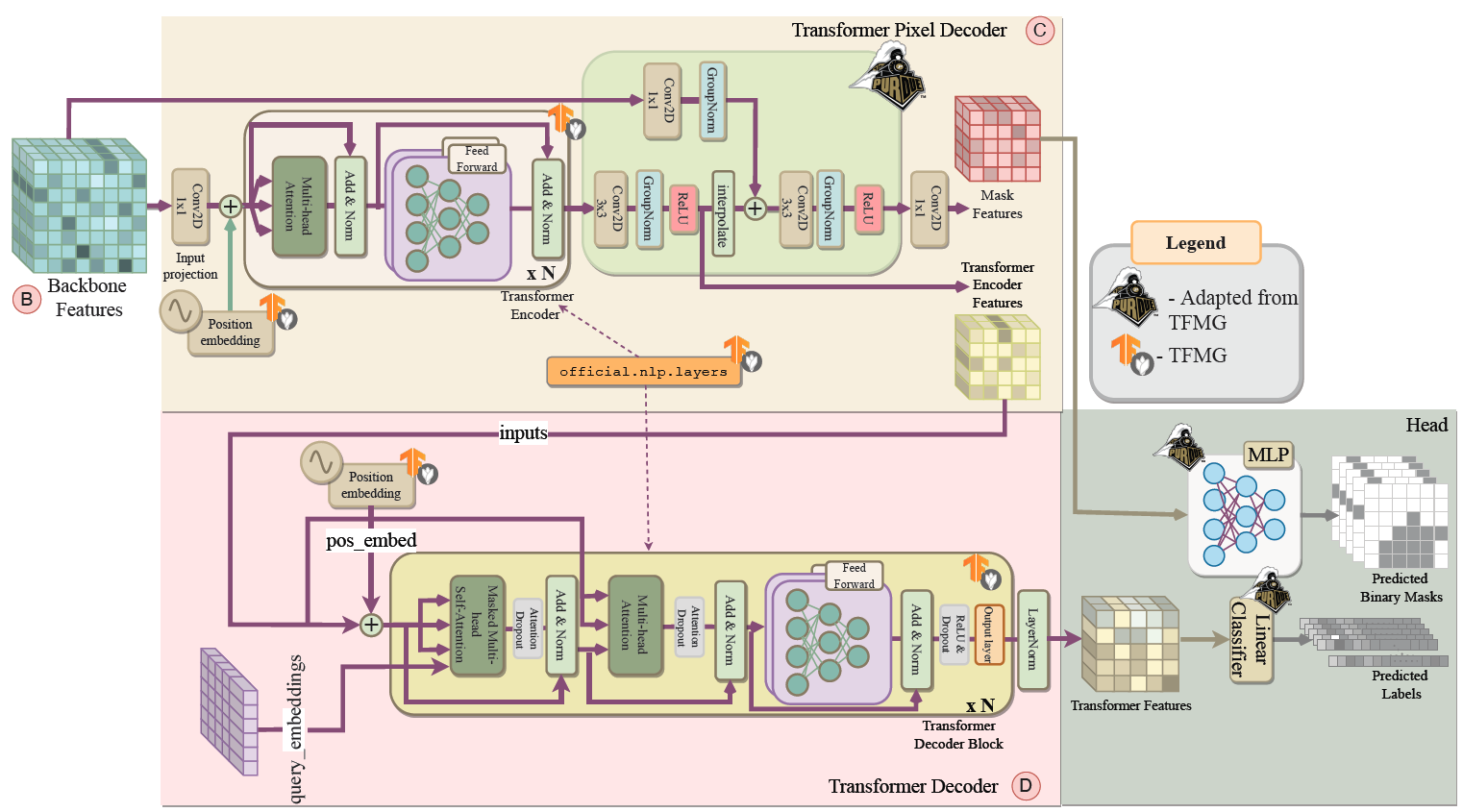}
    \caption{Detailed implementation overview of decoding stage of MaskFormer model. The architecture begins with the backbone features extracted from the input image. These features are then projected and enriched with positional embeddings before being passed through multiple layers of a Transformer encoder. The resulting encoded features are processed by a Transformer Pixel Decoder, which combines them with mask features to enhance localization capability. Simultaneously, query embeddings are combined with positional embeddings and processed by the Transformer Decoder to generate predictions. The decoded features are then passed through an MLP head, which classifies each pixel, resulting in the final output comprising predicted binary masks and their associated labels.}
    \label{fig:detailed_arch}
\end{figure}
\subsection{Model Architecture Implementation Overview}
An overview of the MaskFormer architecture is illustrated in Figure \ref{fig:maskfomer_arch_overview}. Key components and functionalities of the MaskFormer architecture are outlined as follows:
\begin{itemize}
    \item \textbf{Backbone Network \tikz[baseline=(char.base)]{
            \node[shape=circle,draw=red!50,inner sep=1pt, fill=red!50, drop shadow] (char) {A};} :}  The original work explored a diverse array of backbone networks for feature extraction, encompassing both convolutional architectures, such as ResNet-50~\cite{he2016deep}, and transformer-based models like the Swin Transformer~\cite{liu2021swin}. The ResNet50 implementation is readily available in TFMG under \codefont{official > vision > modeling > backbones} and the pre-trained checkpoint for ResNet are made available at TF-Vision Model Garden found under \codefont{official > vision}.
   \item \textbf{Multiscale Features \tikz[baseline=(char.base)]{
            \node[shape=circle,draw=red!50,inner sep=1pt, fill=red!50, drop shadow] (char) {B};} :}  In our model's architecture, a key component is the utilization of multiscale features, as indicated by the `B' in Figure \ref{fig:maskfomer_arch_overview}. This design choice allows for flexibility in feature extraction from the backbone network, specifically ResNet. Our implementation supports both the extraction of deep, high-level features from the final layer of ResNet and the rich, intermediate features available from earlier layers. An important consideration when employing the multiscale features is the increased demand for memory during the model training phase. The inclusion of data from multiple layers escalates the volume of information processed, which can strain available memory resources. In our implementation, the use of multi-scale features is controlled by the parameter \codefont{deep\_supervision}.
\end{itemize}
        
\begin{itemize}
    \item \textbf{Transformer Pixel Decoder \tikz[baseline=(char.base)]{
            \node[shape=circle,draw=red!50,inner sep=1pt, fill=red!50, drop shadow] (char) {C};} :} A comprehensive depiction of the transformer pixel decoder's architecture is illustrated in Figure \ref{fig:detailed_arch}. At its core, the transformer pixel decoder is an assembly of two primary components: the transformer encoder and a series of convolutional and normalization layers. We have incorporated the transformer encoder layers into our architecture by utilizing the existing implementation available in the TensorFlow Model Garden, specifically from the directory \codefont{official > nlp > layers}. To maintain consistency and interoperability with models developed in other frameworks, particularly PyTorch, we have undertaken a process to ensure that our adopted transformer encoder layers exhibit functional parity with their PyTorch counterparts. This involves a series of rigorous verification steps, designed to thoroughly compare and validate the behavior and the output shapes of the encoder layers. The detailed steps we followed for verification are described in section \ref{sec:verification}. The remaining set of convolution and normalization layers in our architecture are implemented utilizing the TensorFlow API.

        \item \textbf{Transformer Decoder \tikz[baseline=(char.base)]{
            \node[shape=circle,draw=red!50,inner sep=1pt, fill=red!50, drop shadow] (char) {D};} :} Similar to our approach with the Pixel Decoder, we have adopted the Transformer Decoder component from the pre-existing implementations provided in the TensorFlow Model Garden, specifically sourced from the directory \codefont{official > nlp > layers}. While integrating the Transformer Decoder from the TensorFlow Model Garden, we ensure that it seamlessly interfaces with other components of our model, such as the transformer pixel decoder and the backbone network. This involves aligning input and output dimensions, ensuring compatible data types, and configuring the decoder appropriately to match our specific requirements. We validate and test the functionality of the integrated transformer decoder and the transformer pixel decoder within our architecture. 
            \end{itemize}

        \tipBox{Converting data to bfloat16 for TPUs offers advantages such as reduced memory bandwidth, improved throughput, preservation of numerical range, compatibility with existing models, and native hardware support. This enables efficient utilization of hardware resources, leading to faster computations and improved performance for deep learning tasks.}

        \errorBox{ \textcolor{red}{\textbf{Warning :}}
        Convergence issues can occur when using bfloat16 for normalization layers in Transformers due to reduced precision. This lower precision affects the accuracy of mean and variance calculations, potentially hindering model learning and convergence. Hence, in our implementation, we do not use bfloat16 for normalization layers in the transformer. Below is the training loss plot for MaskFormer architecture with normalization layers using bfloat16 and float32 datatype. Here, the x-axis represents iterations and the y-axis is the total training loss. \\
        \vspace{0.5em}\includegraphics[width=\textwidth]{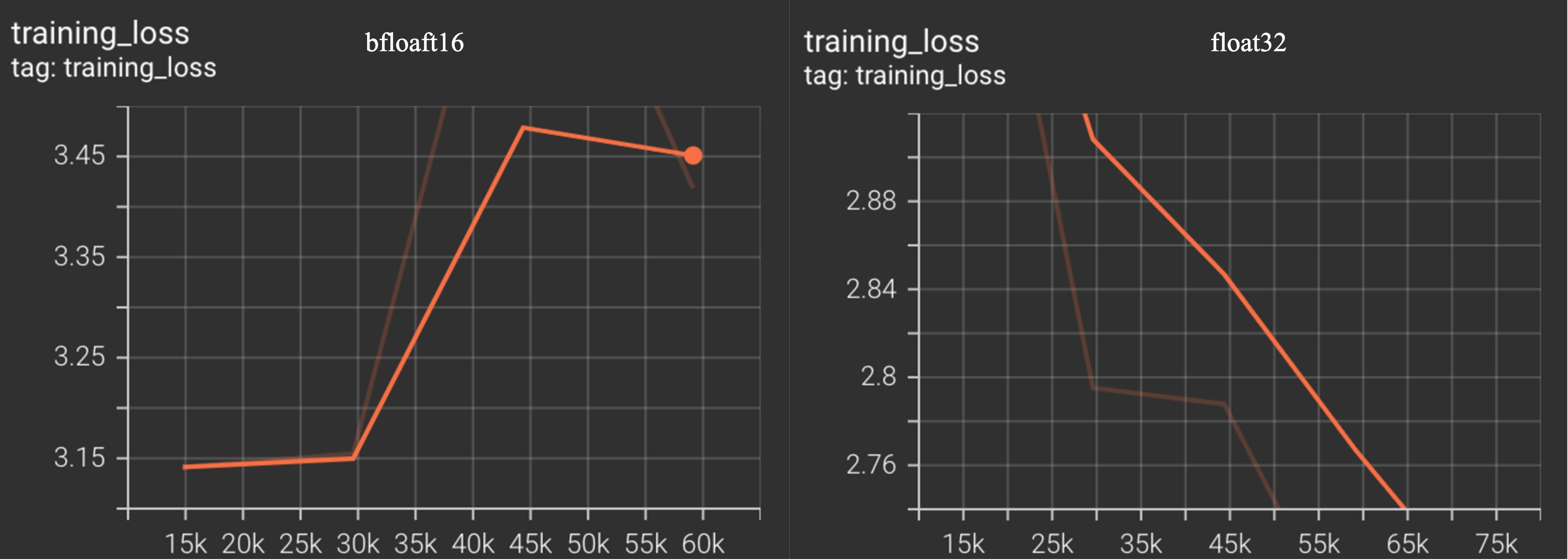}
        }

    \begin{itemize}

    \item \textbf{MLP Projection \tikz[baseline=(char.base)]{
            \node[shape=circle,draw=red!50,inner sep=1pt, fill=red!50, drop shadow] (char) {E};} :} The MLP (Multi-Layer Perceptron) projection layers, in conjunction with the linear classifier, are used to obtain the predicted binary masks and corresponding labels. The Transformer Decoder processes the Transformer Encoder Features to extract high-dimensional features, Transformer Features, that encapsulate both spatial and contextual information are used by the linear classifier to predict the label of each binary mask. Simultaneously, Mask Features that are obtained from the Transformer Pixel Decoder are fed into the MLP projection layers to obtain the predicted binary masks. We implement the Head of MaskFormer using TensorFlow API. 
\end{itemize}
\tipBox{Ensuring consistent reproducibility performance hinges significantly on utilizing appropriate weight initialization across all network layers. In our implementation, we synchronize weight initialization with the original PyTorch implementation. Notably, the prediction head's initialization holds paramount importance due to its sensitivity within the architecture.}

\subsection{Loss Functions}
In this section, we provide in-depth implementation details of various components for the loss calculation process. The function module consists of a matcher and mask losses (focal, dice, and classification loss). The loss calculation first starts with the step of matching between the predicted outputs and the ground truth as there is no guarantee about the prediction order of the masks and labels followed by calculating all the losses.  A detailed description is provided below :
\begin{itemize}

    \item \textbf{Matcher :} Given a set of \(N\) ground truth binary masks for objects in an image, denoted by \(\{\mathbf{M}_{\text{GT}}^{i}, \mathbf{Y}_{\text{GT}}^{i}\}_{i=0}^{N}\), where \(\mathbf{M}^{i}_{\text{GT}}\) and \(\mathbf{Y}_{\text{GT}}^{i}\) represent the \(i^{\text{th}}\) ground truth binary mask and label, respectively. Let the predicted set of binary masks and class labels be denoted by \(\{\mathbf{M}_{\text{pred}}^{i}, \mathbf{Y}_{\text{pred}}^{i}\}_{i=0}^{\text{n\_queries}}\), where \(\text{n\_queries} = 100\) is the fixed number of predictions made by the model. Given the variable number of ground truth masks per image and the fixed number of predicted masks and labels, it is necessary to find the correct order of masks so that the loss calculation for a given ground truth mask is not performed with any arbitrary predicted mask. In the original PyTorch implementation, the matching step was facilitated by the \texttt{scipy.optimize.linear\_sum\_assignment} function, which provided the correct ordering of masks. This function requires a cost matrix as input, which is used to determine the cost of matching a given ground truth with a prediction. When \(N < \text{n\_queries}\), the cost matrix is rectangular. The cost calculation involves evaluating the focal, dice, and classification losses (to be explained in a subsequent section) and weighting them with predetermined weights. The order of the predictions is disregarded at this step. However, directly applying the \texttt{scipy.optimize.linear\_sum\_assignment} function on tensors can significantly reduce training performance on TPUs.
    
    \end{itemize}
    \tipBox{To address the challenge of optimizing the matching process for execution on Tensor Processing Units (TPUs), we leverage the Hungarian Matcher, implementation made available on TFMG. This implementation is adopted from the codebase available at \codefont{official > projects > detr > ops}. Utilizing this optimized implementation significantly accelerates training time by ensuring that all operations required by the matcher are compatible with TPUs, thereby eliminating the need for computational offloading to the CPU.}
   
    \errorBox{\textcolor{red}{\textbf{Warning :}} The output format of the matcher used within the PyTorch framework and that of the TFMG matcher exhibits significant differences, attributable to the distinct methodologies employed by each implementation. This divergence in output formats underscores the necessity for careful consideration when conducting verification processes. Direct one-to-one comparisons of the matcher outputs are not feasible without accounting for these inherent discrepancies.}
    
    \implementationBox{
    In the original PyTorch implementation, a rectangular cost matrix is utilized to facilitate the matching process. However, the matcher available on TFMG requires a square cost matrix for its operations. This constraint necessitates that in our implementation, we adjust the dimensions of the cost matrix to ensure that the number of ground truth masks \( \mathbf{N} \) equals the number of queries \( \text{n\_queries} \), which is fixed at 100. Therefore, to accommodate this requirement and ensure compatibility with the TFMG's matcher, our approach involves equalizing \( \mathbf{N} \) and \( \text{n\_queries} \) to maintain a square cost matrix, via additional padding. Note that additional padding comes with increased memory usage.\\
    The padding must be implemented in such a manner that it does not distort the original cost values or introduce biases in the matching process. We achieve this by assigning the highest cost to the padded indices of the cost matrix. Functionally, the output of matcher used in PyTorch and in TensorFlow is the same despite difference in the shape of cost matrix.}
    
    \begin{itemize}
    \item \textbf{Dice Loss:} Dice Loss is employed to supervise the prediction of masks, offering a measure of similarity between the predicted binary masks and the ground truth masks. It is particularly effective for handling imbalanced datasets where the presence of the class of interest significantly varies across samples. The formulation of Dice Loss is based on the Dice coefficient (also known as the Sørensen-Dice coefficient), which calculates the overlap between two samples. Mathematically, it is defined as:

    \[
    \text{Dice Loss} = 1 - \frac{2 \times | \mathbf{Y}_{\text{pred}} \cap \mathbf{Y}_{\text{GT}} |}{| \mathbf{Y}_{\text{pred}} | + | \mathbf{Y}_{\text{GT}} |}
    \]
    
    where \( \mathbf{Y}_{\text{pred}} \) represents the predicted mask and \( \mathbf{Y}_{\text{GT}} \) represents the ground truth mask. This loss function encourages the model to increase the overlap between the predicted and true masks, thereby improving the accuracy of the predictions for the presence of objects within an image. We implement the dice loss using TF API and ensure that extra padded regions are ignored for calculating the loss.

    \end{itemize}
    \begin{itemize}
    \item \textbf{Focal Loss:} Focal Loss is designed to address class imbalance by modifying the standard Cross-Entropy Loss in such a way that it places more focus on hard, misclassified examples. This is achieved by adding a modulating factor to the Cross-Entropy formula, which decreases the loss for well-classified examples, pushing the model to prioritize the learning of difficult cases. Focal Loss is particularly useful in scenarios where there's a significant imbalance between the foreground and background classes, as often encountered in object detection tasks. The formula for Focal Loss is given by:

\[
\text{Focal Loss} = -\alpha_t (1 - p_t)^\gamma \log(p_t)
\]

where \(p_t\) is the model's estimated probability for the class with label \(t\), \(\alpha_t\) is a weighting factor for the class \(t\), and \(\gamma\) is the focusing parameter that adjusts the rate at which easy examples are down-weighted. We implement the focal loss using TF API and ensure that extra padded regions are ignored for calculating the loss.
\end{itemize}
\errorBox{\textcolor{red}{\textbf{Error :} \codefont{``Unable to find the relevant tensor remote\_handle: Op ID: 30598, Output num: 0", ``grpc\_status":3\} }} \\
        \textbf{Context :} We encountered this error while debugging the loss functions.\\
        \textbf{Description :} The error message indicates a problem in a distributed TensorFlow setup involving gRPC communication.. \\
        \textbf{Solution :} Through our debugging efforts, we discovered that the error arose from employing operations incompatible with XLA. Specifically, operations typically native to Python, such as using \codefont{for} loops for iterative processing and appending elements to a  \codefont{list or dict}, were identified as the root causes. To ensure the code is compatible with XLA we recommend following the best practices provided in TensorFlow Core documentation.}
\begin{itemize}
\item \textbf{Classification Loss:} Classification Loss measures the discrepancy between the predicted class probabilities and the actual class labels of the training examples. We implement the cross entropy loss using TF API and ensure that labels with no objects padded labels are downweighed with a pre-defined weight for calculating the loss.

\end{itemize}

\subsection{Training and Hyper-Parameter Tuning}
In this section, we delve into the convergence challenges that emerged during the model training process, as well as the hyper-parameter tuning that was essential for overcoming these obstacles. The default parameters inherited from the original implementation proved to be inadequate for facilitating training convergence. Some of the critical hyper-parameters that were tuned are enumerated below.

\textbf{Weight Initialization :} The standard layer implementations within TFMG diverge from the original MaskFormer implementation, particularly in terms of weight initialization. Such discrepancies can significantly impact model performance and convergence. In our implementation, we ensured that the weight initialization of all the layers matched the original implementation and observed that the weight initialization for the Head played a critical role.

\textbf{Batch Size :} The batch size hyper-parameter turned out to be one of the critical parameters that needed to be tuned to obtain smooth training convergence. Utilizing a smaller batch size for debugging purposes inadvertently led to a model that overfitted the background class.
 
\errorBox{\textcolor{red}{\textbf{Warning :}} The original PyTorch implementation of MaskFormer demonstrates robustness against overfitting to the background class, maintaining generalization even when trained with very small batch sizes. In contrast, our implementation tends to overfit rapidly under similar conditions. This discrepancy suggests that the PyTorch version may benefit from certain implicit regularization in the dataloader as the inputs are not resized to a constant size.}

\textbf{No object class weight :} The primary purpose of this hyper-parameter is to balance the model's attention between areas with objects and areas without objects. In many images, the vast majority of the image might not contain any objects of interest. Without proper weighting, the model could become biased towards predicting \textit{no object} most of the time, because it might minimize the loss more effectively by doing so. This would be counterproductive to the goal of detecting the often much smaller areas that do contain objects. Hence, this parameter needs to be tuned specifically for each dataset. In the original implementation, a no-object weight of 0.1 was used. However, in our version, employing a weight of 0.1 led the model to overfit towards predicting the absence of objects exclusively, always classifying regions as devoid of objects.

\implementationBox{A crucial aspect of our hyper-parameter optimization involved adjusting the ``No object class weight''. The initial values sourced from the PyTorch codebase did not deliver satisfactory outcomes. We discovered that utilizing a significantly lower value of 0.0001 enhanced the model's ability to generalize across different classes.}

In addition to the previously mentioned hyper-parameters, there are numerous other settings, such as gradient clipping and warm-up phases, that can be adjusted to potentially enhance model performance further. Nonetheless, our experience indicated that the three highlighted hyper-parameters were the most crucial for ensuring the model's generalization capabilities. Due to the constraints of this reproducibility project and compute, we were unable to conduct a more in-depth exploration of the hyper-parameters.

\section{Verification}\label{sec:verification}
Previous research has highlighted effective strategies for reengineering deep learning models, as noted in works by~\cite{Jiang2023CVReengineering, Vishnu2021TechReport}.  We adopted these practices for our replication process. In this section, we will provide a detailed, step-by-step account of the procedures we adhered to during our reimplementation efforts.
\subsection{Shape Testing}
Shape testing plays a crucial role in verifying that the data flowing through the model's layers conforms to expected dimensions. Specifically, in models like MaskFormer where multiple modules are interacting with each other, the tensor shapes must be compatible across modules. Given below are the systematic steps we used to verify the shapes of the tensors. 
\begin{itemize}
    \item \textbf{Input Shape Validation:} We started by validating the shape of the input data. We ensured that at various stages, from converting the raw dataset to TFRecords to processing in the dataloader before and after padding, the final output of the dataloader matched the input shape required by the MaskFormer model. This step helped identify any preprocessing errors or mismatches in data augmentation procedures.
\end{itemize}
\errorBox{\textcolor{red}{\textbf{Warning :}} While shape validation provides a valuable mechanism for verifying the integrity of a module's inputs and outputs, it does not guarantee the absence of errors. For example, the usage of normalization layers does not necessarily alter the shape of tensors however further testing is necessary to find the presence of additional layers.}
\begin{itemize}
    \item \textbf{Layer-wise Output Shape Verification:} For each layer in the model, we programmatically checked the output shapes by feeding a randomly initialized tensor. This verification process is essential to confirm that the transformations applied by each layer (such as convolutions, pooling, transformer layers, and fully connected layers) yield outputs of the correct dimensions. It also helps in identifying layers where shape mismatches occur, facilitating quicker debugging.
\end{itemize}

\errorBox{\textcolor{red}{\textbf{Warning :}} When performing shape verification for inputs as well as layer outputs, it is crucial to note the difference in the tensor formats used by TensorFlow and PyTorch. In PyTorch, tensor shapes follow the NCHW format, while in TensorFlow, they follow the NHWC format, where N, C, H, W represent the batch dimension, channels, height, and width, respectively.}

\begin{itemize}
    \item \textbf{Final Output Shape Check:} The last step in shape testing involved validating the shape of the model's final output against the expected dimensions. This ensures that the model's predictions or classifications can be correctly interpreted and applied to downstream tasks.
\end{itemize}

\subsection{Unit Testing}
Unit testing involves testing individual sub-components or components of the MaskFormer to ensure that the implementation is bug-free and functionally equivalent to the PyTorch code.
\begin{itemize}
    \item \textbf{Functionality Tests for Customized Modules :} While customizing the existing code from TFMG, we incorporated several custom layers and functions. To ensure their independent and correct functionality within the model, we devised a comprehensive suite of unit tests. These tests not only encompass the previously described shape tests, which verify the dimensions of data flowing through the model but also rigorously evaluate the actual outputs against expected results for predefined inputs. For instance, in the case of custom loss functions, our validation process involved feeding the model with carefully selected tensors that have known loss values derived from the original model's computations. We then compared the loss values output by our customized implementation against these benchmarks. To quantify the precision of our custom loss function, we established an acceptance threshold: the computed loss values should exhibit negligible deviation from the expected ones, with differences constrained to within $\leq 0.001$.
    \item \textbf{Gradient Flow and Magnitude Checks :} Ensuring that gradients correctly flow through the model's layers is vital for the training process. We employed unit tests to check gradient computation and propagation, particularly in custom components where automatic differentiation might face complexities.
    \end{itemize}
    \errorBox{\textcolor{red}{\textbf{Warning :}} When conducting gradient flow and magnitude checks, it's imperative not to rely solely on comparing the gradient magnitudes between our implementation and the original PyTorch implementation. This caution is warranted because various factors such as minor initialization discrepancies, layer-specific optimizations, and TPU specific optimizations can introduce randomness into the process. Therefore, direct comparison of gradient magnitudes may not accurately reflect the correctness or effectiveness of our implementation.}
    
    \begin{itemize}
    \item\textbf{Model Component Integration Testing:} Beyond testing individual layers or functions, we also conducted tests on how these components integrate. For instance, testing the integration of newly introduced layers with existing ones or verifying the compatibility of custom loss functions with the model's output.
    \item \textbf{Error Handling and Edge Cases:} Finally, we wrote unit tests to cover error handling and edge cases. This includes testing the model's behavior when provided with invalid input types, sizes, or values and ensuring that the model fails gracefully or throws meaningful errors.
\end{itemize}

\begin{table*}[]
\centering
\caption{Comparison of module shapes and mean outputs between original PyTorch and our TensorFlow implementations. ($\dagger$) The table particularly highlights the differences in the output values of the Transformer Decoder module (first row in table). There, the mean output in TensorFlow is notably different due to the inclusion of an additional convolutional layer that is not present in the PyTorch implementation. We use randomly initialized tensors with appropriate shape for shape testing.}
\resizebox{\textwidth}{!}{%
\begin{tabular}{@{}lllll@{}}
\toprule
\textbf{Module} & \textbf{Torch Shape}    & \textbf{TF Shape}        & \textbf{Mean Output (Torch)} & \textbf{Mean Output (TF)} \\ \midrule
Transformer Decoder $\dagger$    & [1, 100, 256]           & [1, 100, 256]            & -0.0026                      & -1.5832484e-09 \\
Transformer Pixel Decoder             & [1, 256, 160, 160]      & [1, 160, 160, 256]       & 0.0542                       & 0.011979061\\
PosEmbed        & [1, 256, 20, 20]        & [1, 20, 20, 256]       & 0.4937                       & 0.49366885 \\
MLPHead         & N/A                        & N/A                     & [0.012966,1.593415]         & [0.012966, 1.593414] \\
Classification loss         & N/A                        & N/A                     & 0.06040         & 0.06043 \\ 
Focal loss         & N/A                        & N/A                     & 0.2129         & 0.21077 \\ 
Dice loss         & N/A       & N/A                     & 0.2980         & 0.29806\\ \bottomrule
\end{tabular}
}

\label{tab:torch_tf_comparison}
\end{table*}

\subsection{Differential Testing}
\paragraph{Weight loading}
To conduct differential testing on our architecture, it is imperative to first import the pre-trained weights. 
However, unlike previous replication efforts~\cite{Vishnu2021TechReport}, we encountered a significant obstacle: the absence of an open-source TensorFlow checkpoint for MaskFormer. The absence of directly compatible pre-trained weights complicates differential testing on the model, necessitating manual framework-to-framework conversion. This process is time-consuming and requires meticulous verification to ensure (1) successful transfer of weights for each layer, and (2) correctness of all converted weights.

In an initial attempt to bridge this gap, we explored the possibility of converting MaskFormer's pre-trained weights, hosted on its GitHub repository, into the \textit{Open Neural Network Exchange} (ONNX) format~\cite{ONNX}.
ONNX is designed as an open format to represent machine learning models and facilitates interoperability between different frameworks, including PyTorch and TensorFlow~\cite{jajal2023analysis}. Theoretically, this approach should have allowed us to leverage open-source model converters, such as \textit{torch2onnx} and \textit{onnx-tf}, to transition the model's checkpoints from the Torch format to TensorFlow. The conversion process involves loading the .\textit{onnx} model and converting it into a .\textit{pb} TensorFlow graph, which is subsequently loaded into \textit{tensorflow}. This process generally preserves the structure of the model but does not facilitate direct manipulation or extraction of weights as individual tensor objects. Instead, the converted static graph, running on the \textit{tf-backend}, encapsulates both the architecture and the weights but prevents the transfer of individual layer weights to an existing TensorFlow model as a separate entity.

To surmount these challenges, we developed a custom converter tool capable of directly translating the weights from PyTorch to TensorFlow. This manual conversion process involved intricate mapping and adaptation of model parameters, ensuring compatibility and functional integrity of the architecture in its new framework. Our tool not only addresses the immediate issue of importing pre-trained weights but also contributes to the broader field by offering a potential workaround for similar conversion challenges involving complex models with framework-specific operations. A summary of the differential testing output comparison between our implementation and the original PyTorch implementation is shown in Table \ref{tab:torch_tf_comparison}.
Our custom converter tool is available at \href{https://gist.github.com/AkshathRaghav/9f81ea6f997a2972732fb3d955b5b444}{this link.}

\section{Debugging and Performance Evaluation}
In this section, we outline the debugging setup used for the development of MaskFormer, as well as a qualitative assessment of the trained model. Owing to computational constraints, our evaluation is preliminary but presents substantial proof of the efficacy of our implementation. It indicates that with extended training time and hyper-parameter optimization, our implementation has the potential for significant improvements.
\begin{figure}
    \centering
    \includegraphics[width=\textwidth]{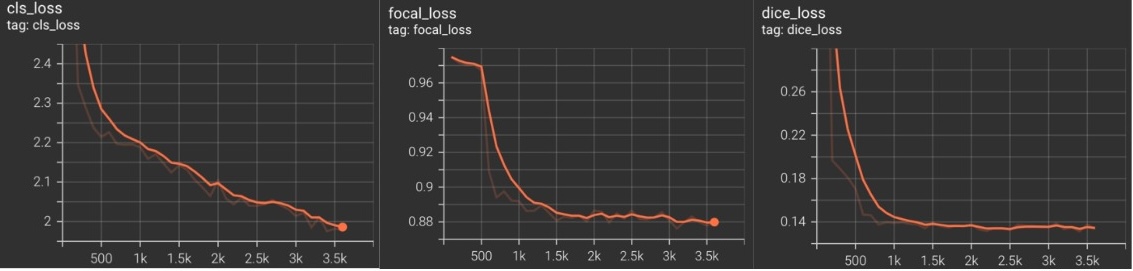}
    \caption{Loss Curves Visualization on TPUs using TensorBoard: raw loss values (light color) and smoothed values (dark color).}
    \label{fig:debug_losses_dry_run}
\end{figure}
\subsection{Debugging Setup and Tools}\label{sec:debugging}
The setups we used for debugging and training our model are as below:
\begin{itemize}
\item \textbf{Debugging:} Our debugging setup to dry run and test various components of MaskFormer consisted of an AMD EPYC 7543 32-core Processor and a single NVIDIA A100 (80GB) GPU. We utilized TensorFlow's eager execution mode for line-by-line debugging. Additionally, we employed logging via TensorBoard for real-time monitoring of the loss function, gradients, and activations. Additionally, one can also debug the model using CPU, however, it would be very slow and will require a significant amount of RAM. The dry runs on GPU are limited to the input image size of 640 x 640 with a batch size of 1.

\item \textbf{Training :} Our model training infrastructure uses Cloud TPUs provided by Google Cloud Platform (GCP). Specifically, we use TPU node architecture. The architecture of the TPU Node is designed such that the user VM interacts with the TPU host using gRPC for communication. In this setup, direct access to the TPU Host is not possible, which can complicate the process of debugging training sessions and resolving TPU-related issues. However, several visualization tools are made available by GCP to monitor various aspects of the training and debugging process. Some useful monitoring tools that we found useful include as detailed below.

The Google Cloud Platform (GCP) offers a comprehensive suite of monitoring tools designed to track the performance and health of Cloud TPUs. Using Google Cloud Monitoring, you can automatically collect metrics and logs from both the Cloud TPU and its associated Compute Engine host.
\begin{figure}
    \centering
    \includegraphics[width=\textwidth]{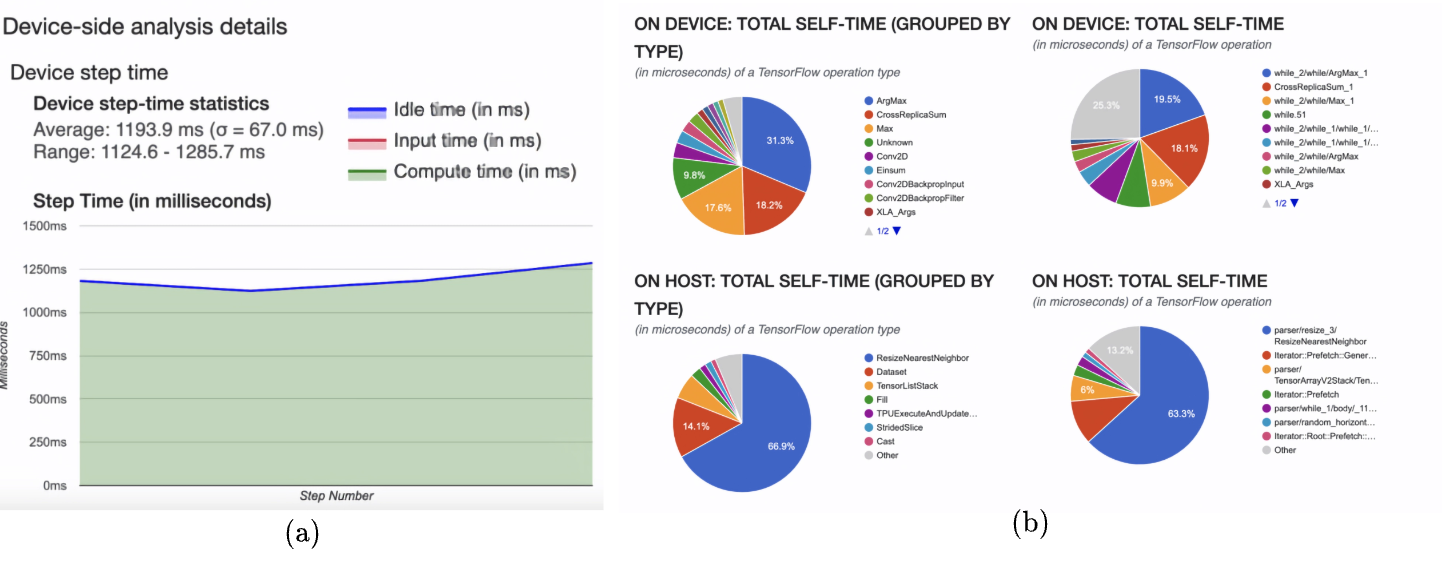}
    \caption{TPU Profiling Results. Subfigure (a) displays the device step time breakdown, while subfigure (b) illustrates the time allocation for different operations on the device and the host. Using the time spent on operations shown in  subfigure (b) one can identify specific costly operation like \codefont{ArgMax} operation on TPU and optimize our implementation.}
    \label{fig:tpu_profile}
\end{figure}
\section*{Key Monitoring Tools and Features in GCP}

\begin{itemize}
    \item \textbf{Metrics Collection:} Automatic tracking of numerical quantities over time, such as CPU utilization, network traffic, and Tensor Cores' idle durations. These metrics are crucial for gauging the performance and resource consumption of Cloud TPUs.
    
    \item \textbf{Logging Capabilities:} Logs are essential for documenting events at precise moments. Generated by Google Cloud services, third-party applications, or custom code, these log entries are instrumental in troubleshooting. Creating log-based metrics from these entries allows for transforming log data into actionable insights.
    
    \item \textbf{Viewing and Querying Data:} The Metrics Explorer in Cloud Monitoring allows for the real-time visualization of metrics. Advanced queries can be made using HTTP calls or the Monitoring Query Language for in-depth data analysis.
    
    \item \textbf{Prerequisites and Setup:} Initial requirements include a Compute Engine VM and Cloud TPU resources, alongside a basic understanding of Google Cloud Monitoring.
    
    \item \textbf{Detailed Metrics for TPUs:} Specific metrics provided for TPUs include \texttt{memory/usage} for monitoring memory usage, \texttt{network/received\_bytes\_count} and \texttt{network/sent\_bytes\_count} for network traffic, \texttt{cpu/utilization} for CPU load, and \texttt{tpu/tensorcore/idle\_duration} for tracking Tensor Cores' activity.
    \end{itemize}
    \section*{TPU Profiling}
    Profiling the model on Cloud TPU Nodes with TensorBoard and the Cloud TPU TensorBoard plugin is critical for optimizing training performance. This involves capturing profiles either via the TensorBoard UI or programmatically, which then allows for a deep dive into performance metrics.
    \begin{itemize}
    \item \textbf{TPU Utilization:} Measures how effectively the TPU resources are being used.
\item \textbf{CPU Utilization:} Indicates the processing load on the CPU.
\item \textbf{Memory Usage:} Tracks the memory consumption of both the TPU and the host machine.
\item \textbf{Infeed/Outfeed Rates:} Analyze the data transfer rates to and from the TPU, identifying potential bottlenecks.
\item \textbf{TensorFlow Operations Time: }Highlights the most time-consuming TensorFlow operations.
\item \textbf{Step Time:} Average duration of each training step, providing insights into overall model efficiency.
\end{itemize}
For further guidance on capturing and analyzing these metrics, consult Google Cloud's documentation on TPU profiling tools.
\item \textbf{Data Storage :} We store the TFRecords in a GCP bucket, located in the same zone as our TPU nodes to enhance data access speed and efficiency. This setup minimizes latency and maximizes throughput.
\end{itemize}

\begin{figure}
    \centering
    \includegraphics[width=\textwidth]{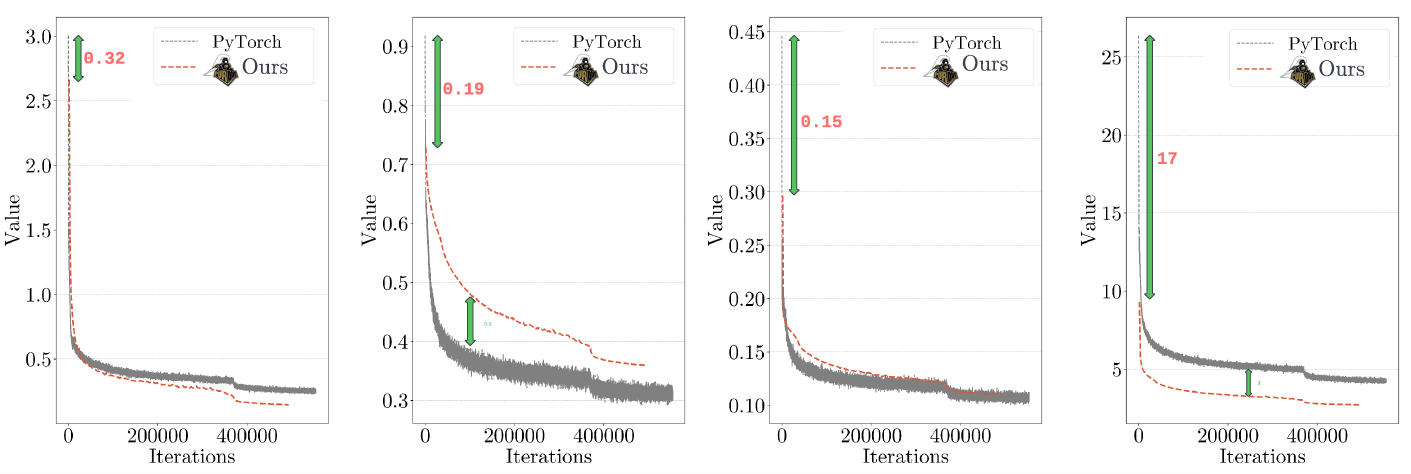}
    \caption{Comparative convergence plots of loss values over training iterations for models implemented in PyTorch and TensorFlow. From left to right: Classification Loss, Dice Loss, Focal Loss, and Total Loss.}
    \label{fig:final_loss_plots}
\end{figure}
\subsection{Qualitative Results}
We organize the qualitative evaluation of our model into two distinct sections:

\subsection{Training Performance Analysis}
This section delves into the performance of the model during the training phase. We assess how well the model learns and adapts to the training dataset, which is crucial for its ability to generalize from the training examples provided. Utilizing the TPU profiling and monitoring tools, we examine a variety of metrics, such as CPU and TPU utilization, memory usage, and data transfer rates, as detailed in section \ref{sec:debugging}. This analysis helps us identify potential bottlenecks in our training process and informs decisions on model adjustments and optimizations to enhance learning efficacy.

\textbf{TPU Profiling Results :} During the development of MaskFormer, we periodically leveraged the TPU profiling tool to assess bottlenecks in the training pipeline. The TPU profiling results, as depicted in Figure \ref{fig:tpu_profile}, provided us with invaluable insights into the efficiency of our model's computations. Figure \ref{fig:tpu_profile}(a) illustrates the device step time, which includes the average, range, and standard deviation of time taken for each step in milliseconds. The step time is further broken down into idle time and compute time, indicating periods when the TPU is not actively processing. This granularity allowed us to identify and minimize idle times by optimizing the input pipeline and ensuring a consistent feed of data into the TPU, thus reducing the time spent waiting for data.

Figure \ref{fig:tpu_profile}(b) presents a comprehensive breakdown of the self-time spent on different types of operations both on the device and the host. The pie charts demonstrate the proportion of time consumed by various TensorFlow operations. For instance, on-device, we observed that a significant portion of time was devoted to matrix multiplication operations, which are computationally intensive but crucial for deep learning tasks. Meanwhile, the host spent a considerable amount of time on data preprocessing tasks. By analyzing these segments, we were able to fine-tune our model's architecture and the data preprocessing steps to better balance the workload between the host and the TPU.

The profiling information guided us in optimizing our training loop. Adjustments made from these insights included tweaking batch sizes, streamlining TensorFlow operations to reduce complexity, and revising the scheduling of operations to avoid underutilization of the TPU. 
\begin{figure}
    \centering
    \includegraphics[width=0.75\textwidth]{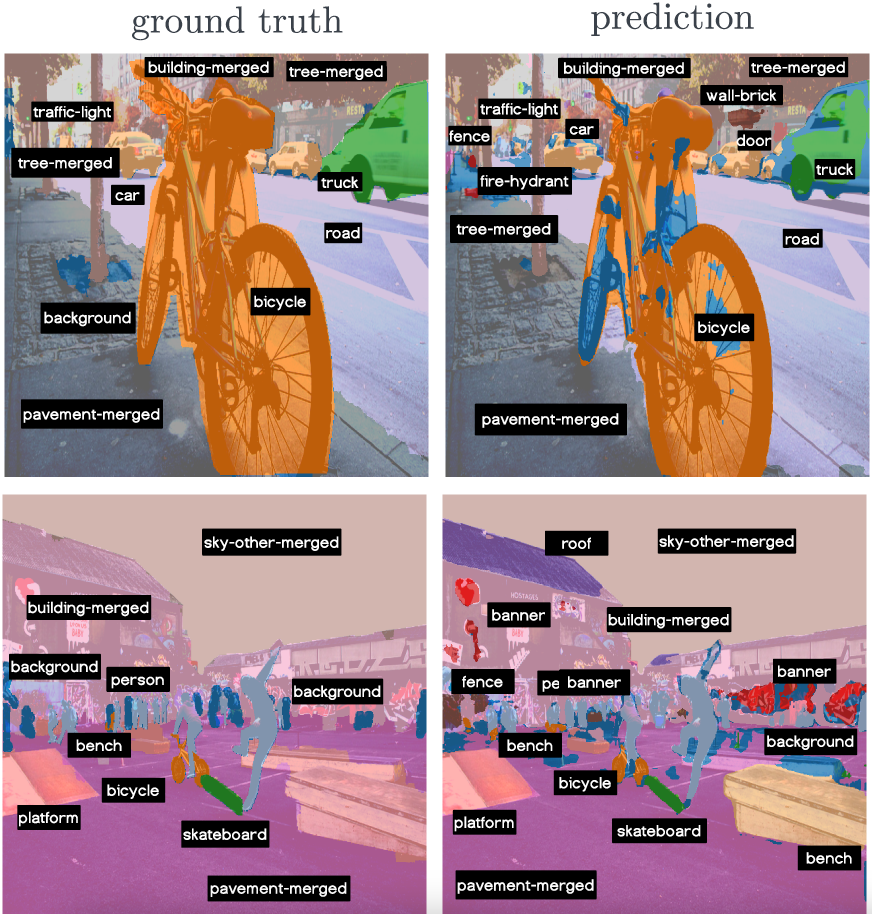}
    \caption{Side-by-Side Comparison of Ground Truth and Model Predictions.}
    \label{fig:results_1}
\end{figure}
\begin{figure}
    \centering
    \includegraphics[width=0.75\textwidth]{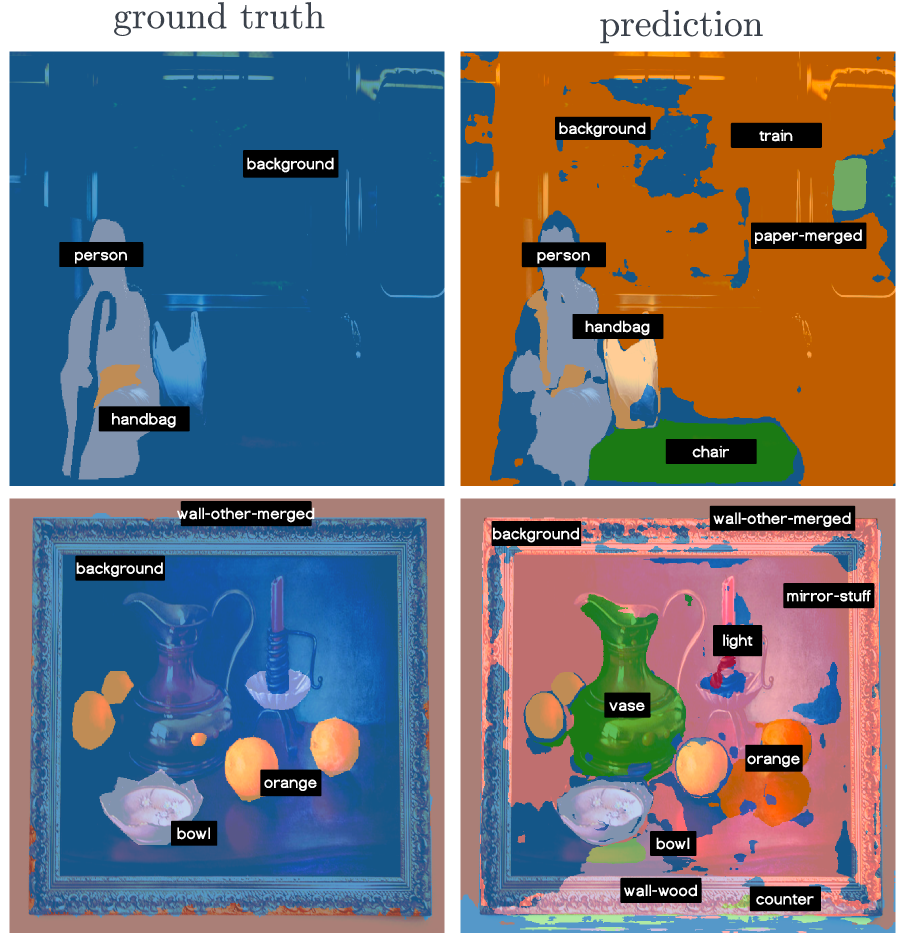}
    \caption{Side-by-Side Comparison of Ground Truth and Model Predictions.}
    \label{fig:results_2}
\end{figure}
\subsection{Evaluation on the Test Set}
Due to the limited scope of the project and constraints in computational resources, this section focuses exclusively on qualitative results obtained from the test set images. We provide a detailed visual examination of the model's output, comparing it against ground truth.

Our evaluation begins with an examination of parameter count and FLOPs presented in Table \ref{tab:comparision_table} and a dry run of training with the loss curves visualized on TensorBoard, as presented in Figure \ref{fig:debug_losses_dry_run}. The figure shows the trajectories of three key loss components during training: classification loss, focal loss, and dice loss. Notably, all three loss components exhibit a downward trend, indicating successful learning. Furthermore, Figure \ref{fig:final_loss_plots} offers a comparative analysis of convergence plots, contrasting our model's loss values over training iterations against the original model implemented in PyTorch. The plots illustrate the classification, dice, and focal losses, as well as the total loss. These graphs are particularly insightful as they show our model, denoted by the red dashed line, consistently achieving lower loss values earlier in the training process compared to the PyTorch implementation. This suggests that our model implemented in TensorFlow with TPU optimization converges.

Lastly, we assess the model's performance through a direct visual comparison of predictions with ground truth, as illustrated in Figure \ref{fig:results_1} and \ref{fig:results_2}. The side-by-side images display how the model segments various objects within urban street scenes. While the model demonstrates strong alignment with the ground truth in many cases, particularly in distinguishing between different categories of vehicles and urban infrastructure, there are discrepancies worth noting. For instance, the model occasionally misclassifies similar textures or confuses overlapping objects, challenges that are common in semantic segmentation tasks. These observations are instrumental in pinpointing areas where the model could be further refined to enhance its segmentation accuracy.

\begin{table}[]
\caption{Comparative Overview of DETR and MaskFormer Implementations. The table summarizes the differences in backbone structure, input size flexibility, and parameter count, between MaskFormer models implemented in PyTorch, and our MaskFormer implementation in TensorFlow.}
\resizebox{\textwidth}{!}{%
\begin{tabular}{llll}
\hline
{\textbf{Implementation}} & \textbf{Backbone} & \textbf{Input size} & \textbf{\# of parameters} \\ \hline
MaskFormer-PyTorch    & ResNet50 + 6 Encoder & Variable size & 45.04M \\ 
MaskFormer-TensorFlow (ours) & ResNet50 + 6 Encoder  & 640 x 640  & 45.16M \\ \hline
\end{tabular}%
}
\label{tab:comparision_table}
\end{table}

\section{Discussion}
\subsection{Project Cost}\label{disc-projectCost}

\begin{table}[h]
    \caption{Google Cloud Platform (GCP) Cost Report.}
    \centering
    \begin{tabular}{lll}
        \toprule
         \textbf{GCP Service} & \textbf{Details} & \textbf{Cost} \\
         \midrule
         Compute engine & Virtual machine instance & \$4811.69 \\
         Cloud storage & Standard storage, data transfer & \$796.89 \\
         Cloud logging & Log storage cost & \$104.28 \\
         Networking & Network intelligence center resource hours & \$23.33 \\
         \bottomrule
    \end{tabular}
    \label{tab:cost}
\end{table}

We provide an estimate of the project's cost in terms of both time and money for reference.
Table~\ref{tab:cost} details the cost of Google Cloud Platform (GCP) resources.
This cost omits the TPU time, to which Google gave us access for the project duration.
During the one year of the project, we attempted two distinct organizations.
Initially, two Ph.D. students led a team of ten undergraduate students for the first six months.
For the remaining six months, the team composition changed to two Ph.D. students and one experienced undergraduate student.

\subsection{Alternative approaches to ML model re-use (``Build vs. buy'')}
Davis \textit{et al.} described several paths to re-using machine learning models~\cite{davis2023JVA}.
This study applied what they call conceptual re-use: given an existing paper and implementation, we developed an independent implementation targeting another development framework and hardware resources.
This approach produces a model tailored to the hardware platform, with maximal adherence to the original paper~\cite{montes2022discrepancies}, and with no known backdoors or other security issues.
However, the cost of a conceptual re-use approach is high (\S \ref{disc-projectCost}). 

Pre-trained models (``adaptation re-use'') present an attractive pathway for model reuse that can significantly reduce the engineering overhead associated with custom optimizations~\cite{jiang2023PTMReuse, jiang2023PTMNaming}. 
The strength of PTMs lies in their versatility and the breadth of their training, which allows for a more straightforward adaptation to a variety of tasks without the need for extensive re-engineering~\cite{han2021pre}. This characteristic of PTMs can accelerate the development cycle, and potentially can also help with the model reproducibility and replicability problem. 
Furthermore, with access to detailed training logs and configurations of open-source PTMs, there is potential for leveraging these resources to assess the accuracy of model replications across various frameworks, illustrating PTMs' broader impact on fostering model consistency and reliability~\cite{Jiang2023CVReengineering}.

The third path discussed by Davis \textit{et al.} was deployment reuse.
Interoperability tools, such as ONNX model converters, facilitate the adaptation of models across different frameworks and hardware~\cite{shen2021comprehensive}.
Despite the potential for performance optimization on such hardware, deployment reuse also has hazards, \textit{e.g.} model conversion errors~\cite{jajal2023analysis,openja2022empirical}.

These cost/benefit trade-offs underscore a key challenge in the reproducibility of AI research. Our findings emphasize the ongoing difficulties associated with training models on TPUs, primarily due to limited framework support and technical complications. We advocate for targeted research efforts to overcome these barriers, enhancing both the deployment efficiency and the replicability of models across diverse computational environments.

\section{Conclusion}
This paper documented our journey and methodologies in replicating the MaskFormer model from its original PyTorch implementation to TensorFlow, specifically optimized for TPUs. Given the limited compute resources and time constraints, our project achieved a partial replication, highlighting the significant engineering challenges and computational demands associated with such undertakings. Throughout this process, we encountered numerous technical hurdles, from framework-specific functionalities and model architecture discrepancies to weight conversion and optimization for TPU acceleration. To address these challenges, we developed custom tools, adapted existing TensorFlow functionalities, and proposed a detailed verification strategy to ensure the fidelity of our replication. Our systematic approach, detailed documentation, and the solutions we proposed aim to serve as a valuable resource for the community. By sharing our experiences and the lessons learned, we hope to contribute to the ongoing dialogue on enhancing reproducibility.

\section*{Acknowledgments}
This research was supported by a gift from Google and by the US National Science Foundation under Grant No. 2107230. We thank these organizations for their support.
\bibliography{main}
\bibliographystyle{tmlr}

\appendix

\end{document}